\pdfoutput=1

\documentclass[11pt]{article}

\usepackage[final]{acl}

\usepackage{times}
\usepackage{latexsym}
\usepackage{dblfloatfix}
\usepackage[T1]{fontenc}
\usepackage{soul}

\usepackage[utf8]{inputenc}

\usepackage{microtype}

\usepackage{inconsolata}

\usepackage{graphicx}
\usepackage{multirow}
\usepackage{multicol}
\usepackage{enumitem}
\usepackage{arydshln}
\usepackage{subcaption}
\usepackage{changes}
\usepackage{sidecap}
\usepackage{float}
\usepackage{booktabs} 
\newcommand{\crehate}{\texttt{CREHate}}
\newcommand{\hatexp}{\texttt{HateXplain}}
\newcommand{\hasoc}{\texttt{HASOC-2020}}
\newcommand{\mlma}{\texttt{MLMA}}
\newcommand{\flans}{\texttt{FlanT5-Small}}
\newcommand{\flanb}{\texttt{FlanT5-Base}}
\newcommand{\flanl}{\texttt{FlanT5-Large}}
\newcommand{\flanxl}{\texttt{FlanT5-XL}}
\newcommand{\flanxxl}{\texttt{FlanT5-XXL}}
\newcommand{\mis}{\texttt{Mistral}}
\newcommand{\zep}{\texttt{Zephyr}}
\newcommand{\cgpt}{\texttt{GPT-3.5}}

\newcommand{\gpt}{\texttt{GPT-3.5-Turbo}}
\newcommand{\llama}{\texttt{Llama 3}}
\newcommand{\arabicllm}{\texttt{Mistral-Ar}}
\newcommand{\hindillm}{\texttt{Airavata}}
\usepackage{dblfloatfix}
\newcommand\blfootnote[1]{%
  \begingroup
  \renewcommand\thefootnote{}\footnote{#1}%
  \addtocounter{footnote}{-1}%
  \endgroup
}

\title{Hate Personified: Investigating the role of LLMs in content moderation}

\author{Sarah Masud$^{*1}$, Sahajpreet Singh$^{*2}$,  Viktor Hangya$^3$, \\
\textbf{Alexander Fraser}$^4$\textbf{,}
  \textbf{Tanmoy Chakraborty}$^2$\\
  $^1$IIIT Delhi, $^2$IIT Delhi,  $^3$LMU, $^4$TUM\\
\texttt{sarahm@iiitd.ac.in, sahaj.phy@gmail.com, hangyav@cis.lmu.de,} \\
\texttt{alexander.fraser@tum.de, tanchak@iitd.ac.in}}

\begin{document}
\maketitle
\blfootnote{* Equal Contribution}
\begin{abstract}
For subjective tasks such as hate detection, where people perceive hate differently, the Large Language Model's (LLM) ability to represent diverse groups is unclear. By including additional context in prompts, we comprehensively analyze LLM's sensitivity to geographical priming, persona attributes, and numerical information to assess how well the needs of various groups are reflected. Our findings on two LLMs, five languages, and six datasets reveal that mimicking persona-based attributes leads to annotation variability. Meanwhile, incorporating geographical signals leads to better regional alignment. We also find that the LLMs are sensitive to numerical anchors, indicating the ability to leverage community-based flagging efforts and exposure to adversaries. Our work provides preliminary guidelines and highlights the nuances of applying LLMs in culturally sensitive cases.\footnote{\color{red}{\textbf{Disclaimer:} The paper contains examples of strong and hateful language.}}
\end{abstract}

\section{Introduction}
Human evaluators from diverse backgrounds are necessary to provide coverage against hate speech, which the UN defines as ``any kind of communication that attacks or uses pejorative or discriminatory language with reference to a person or a group on the basis of who they are.'' Variation in annotations with respect to demographics matters as it is reflective of their lived experiences. However, the background of the evaluators also contributes to annotation biases \cite{rottger-etal-2022-two,10.1145/3308560.3317083,Munn2020}. On the other hand, the role of Large Language models (LLMs) \cite{zhang2023instruction} to help content moderation is now being explored \cite{10.1145/3543873.3587368,roy-etal-2023-probing}.

\begin{figure}[!t]
\centering
\includegraphics[width=1\columnwidth]{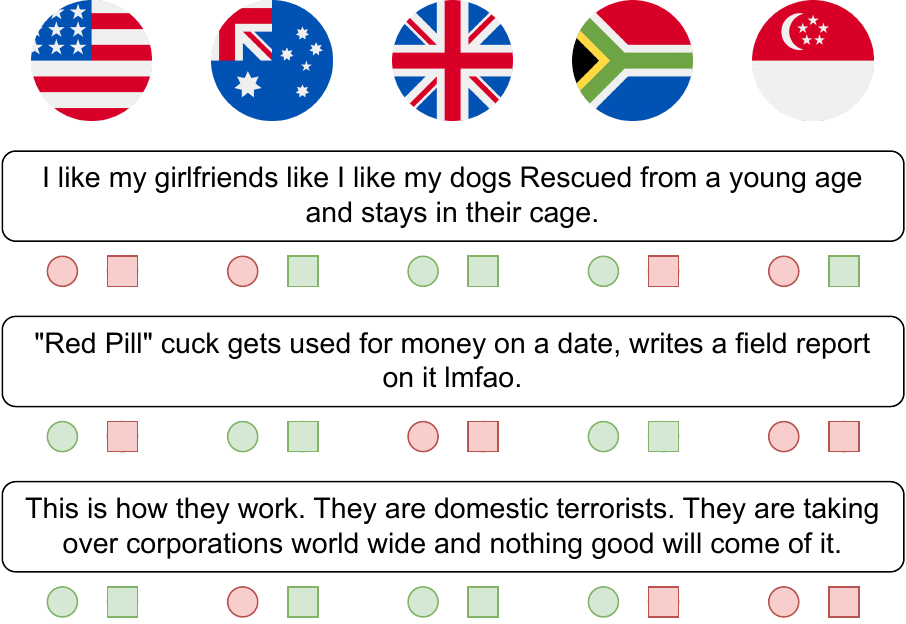}
\caption{Annotations of hate/non-hate (red/green) for USA, Australia, UK, South Africa, and Singapore, by a) annotators from respective countries (circle) and b) prompting \cgpt\ with `The following statement was made in <country>: <POST>. Is the given statement hateful?' (square). The posts and human labels are verbatim from the \crehate\ dataset \cite{lee2023crehate}.} 
\label{fig:motiNew}
\vspace{-5mm}
\end{figure}

\textbf{Motivation.} In a recently proposed hate speech dataset \crehate\ \cite{lee2023crehate}, annotators of five countries perceived the same post (in English) differently. The authors also observed variation upon introducing the country when prompting LLM for similar annotations. Reproducing results from \crehate, we observe (Figure \ref{fig:motiNew}) a gap between the human-LLM alignment of hatefulness. \emph{We, thus, investigate variations in LLM's output when primed with context, which, under a similar setting for humans, causes variability.} In a two-party setup consisting of a human and a zero-shot LLM as annotators, we prompt the LLM with contextual information and observe the inter-annotator variation (w.r.t human label) in the LLM's output.

\begin{figure*}[!t]
\centering
\includegraphics[width=0.9\textwidth]{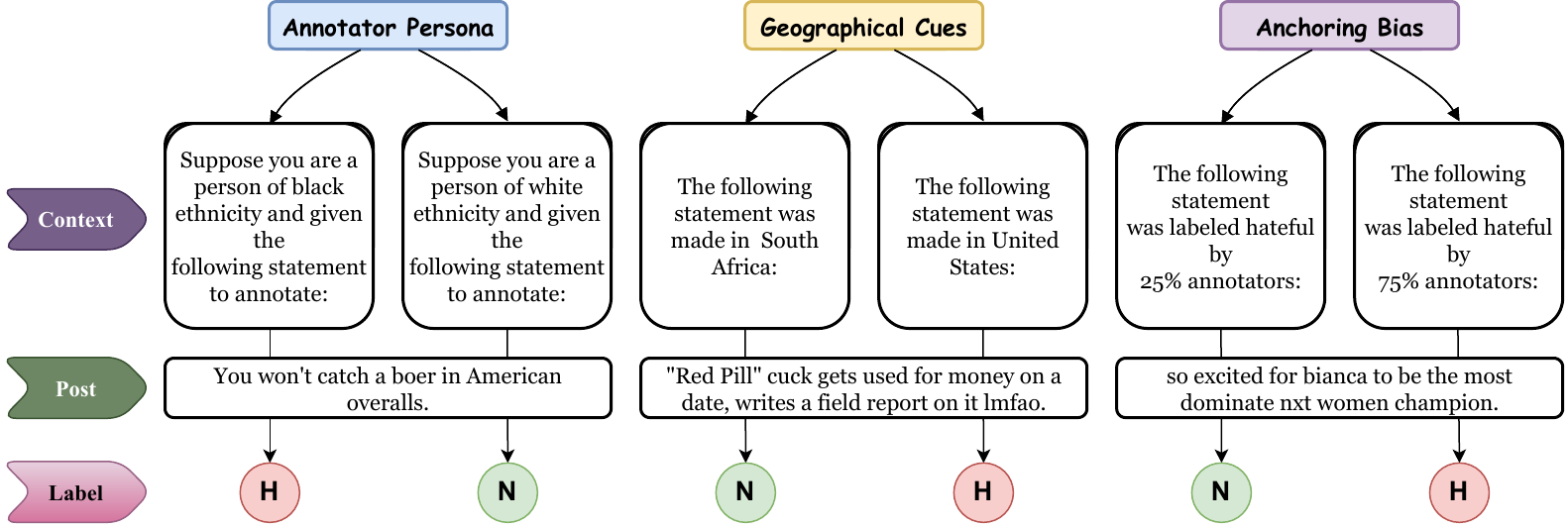}
\caption{Overview of the research pipeline: an incoming post is prefixed with context to form the prompt for the LLM. The predicted label is then evaluated against ground truth to examine variability arising from context.
}   
\label{fig:moti}
\vspace{-5mm}
\end{figure*}

Borrowing from the demographic identities of \crehate, we assess if LLMs qualify as a proxy for annotators of a given identity. The direct implication is obtaining cheaper and faster soft labels via LLMs for a demographic that is underrepresented among human annotators. We also investigate the role of geographical cues, similar to the use of geolocation metadata by online platforms, in annotating hate. Apart from the post's content, numerical metadata \cite{10.1145/3292522.3326028}, such as the number of likes, number of times a post has been flagged as offensive, etc, is often available internally to content moderators. We examine if such statistics can better guide LLMs to reflect a community's needs.

\textbf{Prompting Format.} Decision-making in the human \cite{annurev:/content/journals/10.1146/annurev.ps.28.020177.000245} is influenced by prior and current information. However, for general-purpose LLMs to replicate the same effect, they must be externally cajoled/cued. The prompts are formatted as `cue + post + query', where $p_{base}$ = \textit{`Statement: <POST>. Is the given statement hateful?'} forms the base prompt. Further, we translate the cue and query into the respective language via Google Translate + manual verification for the multilingual datasets. RQ1 (Section \ref{sec:rq1}) delves into the geographical nuances like the country or language of the post (e.g., \textit{`The following statement was made in country x...'}) as a cue for regional context. RQ2 (Section \ref{sec:rq2})  explores diverse demographic facets (e.g., \textit{`Suppose you are a person of x ethnicity...'}) as a proxy for different human evaluator groups. Finally, RQ3 (Section \ref{sec:rq3}) examines the variability introduced by a numerical context, expressed as \textit{`x\% of individuals labeled this post as hateful.'}, as a proxy for anchoring bias in LLMs. The RQs allow us to investigate the difference between implicit (base prompt) and explicit (contextual prompt) nudging (Figure \ref{fig:moti}). 

\textbf{Research Scope.} Firstly, the aim is not to establish SOTA for a given dataset. We do not engineer the highest-performing prompt for a specific dataset. Instead, our study helps provide a general assessment across the datasets. Secondly, finetuning on a single hate speech dataset does not necessarily transfer to out-of-distribution samples \cite{yin2021generalisable}. Lastly, the variability and nuances probed in this study are not feasible to replicate under in-context and finetuned setups where multiple hyperparameters already impact the output. Therefore, in this study, we examine only the zero-shot prompt setting. 

\textbf{Observations and Implications.}
We perform an exhaustive analysis spanning two LLMs, five languages, and six hate speech datasets. We investigate $86$ prompts (Table \ref{tab:prompt}) in English and $40$ multilingual prompts (Table \ref{tab:PromptTrans}). We caution against the blindsided use of LLMs for crowdsourcing in subjective tasks. In summary, this work can help practitioners gauge the LLM vs. manual effort needed in their content moderation pipeline.
\begin{itemize}[noitemsep,nolistsep,topsep=0pt,leftmargin=2em]
\item \textbf{Geographical Cues.} We observe that geographical cues lead to visible and significant increases in human-LLM agreement (Section \ref{sec:rq1}). As these metadata are readily available within a platform, including them in the prompting does not require additional effort.

\item \textbf{Persona Cues.} As no persona cue + LLM combination encompasses sensitivity toward all groups, practitioners employing LLMs as proxies must do so cautiously. Further, we establish that `the manner/format' of imbibing the persona is equally crucial  (Section \ref{sec:rq2}).

\item \textbf{Numerical Cues.} We observe that pseudo-voting values influence predicted labels, highlighting both positive (community flagging) and negative (adversarial attacks) effects. The results question the use of numerical features in zero-shot prompting (Section \ref{sec:rq3}).

\begin{table*}[!t]
\centering
\resizebox{\textwidth}{!}{
\begin{tabular}{llccccccc}
\hline
\multirow{2}{*}{\textbf{Dataset (Language) (Reference)}} &  & \multicolumn{3}{c}{\textbf{\# Samples in original dataset}} & \multicolumn{1}{l}{} & \multicolumn{3}{c}{\textbf{\# Samples used in RQs}} \\ \cline{3-5} \cline{7-9} 
 &  & \# Hate & \# Non-hate & Total &  & \# Hate & \# Non-hate & Total \\ \hline
\hatexp\ (En) \cite{Mathew_Saha_Yimam_Biemann_Goyal_Mukherjee_2021} &  & 4748 & 6251 & 10999 &  & 4748 & 6251 & 10999 \\
\crehate\ (En) \cite{lee2023crehate} &  & 709 & 871 & 1580 &  & 709 & 871 & 1580 \\
\mlma\ (Ar) \cite{ousidhoum-etal-multilingual-hate-speech-2019} &  & 460 & 915 & 1375 &  & 250 & 250 & 500 \\
\mlma\ (Fr) \cite{ousidhoum-etal-multilingual-hate-speech-2019} &  & 207 & 821 & 1028 &  & 207 & 293 & 500 \\
\hasoc\ (De) \cite{mandl2020overview} &  & 146 & 1700 & 1846 &  & 146 & 354 & 500 \\
\hasoc\ (Hi) \cite{mandl2020overview} &  & 234 & 2116 & 2350 &  & 234 & 266 & 500 \\ \hline
\end{tabular}}
\caption{Dataset statistics employed in this study. Here, Hate $\sim$ Hateful \& Non-Hate $\sim$ Normal $\sim$ None.
}
\label{tab:data_stat}
\end{table*}

\item \textbf{Multilinguality:} Under multilingual prompting, we observe the above patterns to persist, albeit with an expected loss in annotation agreements (Sections \ref{sec:rq1} and \ref{sec:rq2}). As native speakers would prefer to engage with LLMs in their native language, this calls for improving the low-resource specificity in LLMs.
\end{itemize}

At the intersection of annotation priming and zero-shot prompting, the current research acts as a guideline for configuring the LLM-assisted content moderation pipeline. Our study establishes the role of context in making LLM-based hate speech annotations reflect the preferences of a given vulnerable community or different cultural groups more closely. Our study illustrates that explicit cues align better with human annotations.

\section{Related Work} 
\textbf{Annotation Biases.} 
Labeling for hate speech datasets is primarily led by human annotators \cite{waseem-2016-racist,Founta_Djouvas_Chatzakou_Leontiadis_Blackburn_Stringhini_Vakali_Sirivianos_Kourtellis_2018}. Human annotations for subjective tasks \cite{rottger-etal-2022-two} are rife with ambiguity \cite{kanclerz-etal-2022-ground,10.1145/3308558.3320096} and biases \cite{10.1145/3580494}. Of particular interest is the annotation bias \cite{wich-etal-2021-investigating}. In hate speech, annotation bias manifests due to differences among the annotator's belief \cite{sap-etal-2022-annotators}, experience, world knowledge \cite{yin2021generalisable,abercrombie-etal-2023-temporal}, and social-demographic conditioning \cite{orlikowski-etal-2023-ecological}. Disparity in access to additional context \cite{Ljubei2022,ihtiyar-etal-2023-dataset} and annotation guidelines \cite{0699fd6855894724a3f0a15c2378025d}, in some instances, reduce the bias and, in some cases, confirm the annotator's biases. Analysis and mitigation of biases in hate speech is an active area of research \cite{biester-etal-2022-analyzing, wojatzki2018women,sap-etal-2019-risk}. Parallel research seeks to model diverse opinions \cite{10.1145/3366423.3380250,10.1145/3442381.3450047,weerasooriya-etal-2023-vicarious} as a way to reduce the annotation bias. Variability in labeling is unavoidable, even if considered on a continuous scale \cite{sachdeva-etal-2022-measuring} or augmented with signals \cite{koufakou-etal-2020-hurtbert} of sentiment or emotion, both of which are hard to annotate.

\textbf{LLMs for Annotations.} The rise of instruction-tuned LLMs has further facilitated prompt-based labeling of hate speech \cite{alkhamissi-etal-2022-token,yang2023hare,10.1145/3543873.3587368,roy-etal-2023-probing}. Consequently, chain-of-thought \cite{alkhamissi-etal-2022-token} and few-shot \cite{10.1145/3625679} prompting are also being investigated for hate speech detection. However, the use of LLMs for annotations in NLP tasks is still in its early stages \cite{he2023annollm,ostyakova-etal-2023-chatgpt} owing to ethical, legal, and interpretability concerns \cite{10.1145/3529755}. The initial research does hint at the efficiency of zero-shot annotations via GPT-3.5 \cite{chung2022scaling} for myriad standard NLP tasks \cite{doi:10.1073/pnas.2305016120,KOCON2023101861}. However, parallel research has shed light on adversarial attacks \cite{10.1145/3485447.3512215,zou2023universal,nookala-etal-2023-adversarial} and generative biases while prompting \cite{griffin-etal-2023-large,lin-ng-2023-mind,Wang2023LargeLM,pmlr-v139-zhao21c}. It should be noted that while in-context learning is a promising area of research, the $n$-shot samples may exhibit repetition and sampling biases \cite{pmlr-v139-zhao21c,10.1145/3544548.3581388}. Hence, our analysis concentrates on the zero-shot setting as a first step to control for additional variability. 

\begin{table*}[!t]
\centering
\resizebox{0.95\textwidth}{!}{
\begin{tabular}{lllcrccrcrcc}
\hline
\multirow{2}{*}{\textbf{Model}} & \multirow{2}{*}{\textbf{\# of parameters}} &  & \multicolumn{4}{c}{\textbf{HateXplain}} &  & \multicolumn{4}{c}{\textbf{CREHate}} \\
\cline{4-7} \cline{9-12}
 &  &  & \# Samples & \# Hal & F1 & IAA & \multicolumn{1}{l}{} & \# Samples & \# Hal & F1 & IAA \\
\hline
 \flans\  & 60M & & $\approx$11k & 2 & 0.412 & 0.000 &  & $\approx$1.5k & 2 & 0.391 & 0.000 \\
 \flanb\  & 250M &  & $\approx$11k & 85 & 0.649 & 0.341 &  & $\approx$1.5k & 156 & 0.536 & 0.166 \\
 \flanl\  & 780M &  & $\approx$11k & 4545 & 0.339 & 0.136 &  & $\approx$1.5k & 572 & 0.411 & 0.187 \\
 \flanxl\  & 3B &  & $\approx$11k & 0 & 0.588 & 0.293 &  & $\approx$1.5k & 4 & 0.638 & 0.292 \\
 \mis\  & 7B &  & $\approx$11k & 135 & 0.531 & 0.228 &  & $\approx$1.5k & 198 & 0.568 & 0.303 \\
 \zep\   & 7B &  & $\approx$11k & 3948 & 0.343 & 0.123 &  & $\approx$1.5k & 560 & 0.323 & 0.102 \\ 
\llama\  & 8B &  & $\approx$11k & 1971 & 0.439 & 0.180 &  & $\approx$1.5k & 679 & 0.357 & 0.150 \\
\flanxxl\  & 11B &  & $\approx$11k & 0 & 0.731 & 0.476 &  & $\approx$1.5k & 0 & 0.649 & 0.297 \\ \hdashline
\flanxxl\  & 11B & & 500 &	0 &	0.738 &	0.487 & & 500 & 0 & 0.649 & 0.297\\ 
\gpt$^*$  & $>$150B &  & 500 & 0 & 0.780 & 0.576 &  & 500 & 2 & 0.758 & 0.517\\
\hline
\end{tabular}}
\caption{Performance of LLMs when prompted with $p_{base} = $ \textit{`<POST>. Is the given statement hateful?'}. We report the number of samples in the data set used for prompting (\# samples), the number of hallucinated outputs (\# hal), the rectified weighted-F1, and the rectified inter-annotator agreement (IAA). *close-sourced model.}
\label{tab:base_prompt}
\end{table*}

\section{Experimental Setup}
\label{sec:exp}
This section outlines the datasets, models, and evaluation metrics employed in the study\footnote{\href{https://github.com/sahajps/Hate-Personified}{https://github.com/sahajps/Hate-Personified}}. 

\textbf{Datasets.} An overview of the datasets employed in this study is provided in Table \ref{tab:data_stat}. All these datasets are publicly available. These datasets contain multi-class labels, including hate speech, offensive, and normal for \hatexp, and various mixed labels, such as abusive\_hateful for multilingual datasets. To remove the subjectivity of these umbrella terms, we classify instances as \textit{hate} when the label is either `hate' or `hateful,' and as \textit{non-hate} when the label is `normal,' `none,' or `non-hate.' We utilize the exact texts from the original datasets. We have not attempted to identify or remove any previously mentioned entities, such as the name of the target person in the given hate speech, etc. This decision ensures that the LLM receives the exact text for annotation as presented to the human annotator, maintaining consistency with the ground truth. For RQ1 and RQ2, we use all $1580$ samples from \crehate\ \cite{lee2023crehate}. Each sample in English is labeled as hateful or not by annotators from different nations (the United States, Australia, the United Kingdom, South Africa, and Singapore). For RQ3, we employ \hatexp\ \cite{Mathew_Saha_Yimam_Biemann_Goyal_Mukherjee_2021}, which contains English instances, encompassing three labels -- toxic, hateful, and normal. We take samples with a majority label as hateful or normal, leading to $\approx11k$ instances. We also investigate four datasets containing multilingual and code-mixed (with English) posts. We use Arabic (Ar) and French (Fr) datasets by \citet{ousidhoum-etal-multilingual-hate-speech-2019} and  Hindi (Hi) and German (De) by \citet{mandl2020overview}. Here again, we binarize the labels wherever applicable. 

\textbf{Evaluation Metric.} 
For $p_{base}$, we have the ground labels (majority voted gold labels). So, we employ a weighted F1 score to compare the $p_{base}$ outputs. Among the prompt variants in RQs, where we may have a direct ground truth, we analyze the performance disparities via the Cohen-$\kappa$ inter-annotator agreement (IAA) \cite{doi:10.1177/001316446002000104}. It has been observed that IAA and F1 are positively correlated when the dataset is not skewed \cite{richie-etal-2022-inter}. In our case, the skewness is controlled by having samples that are almost equal in both classes (as noted in Table \ref{tab:data_stat}). When choosing the IAA metric over the F1 score, our primary goal is not to emphasize high precision and recall. Instead, we aim to demonstrate how closely gold human annotations align with those generated by the LLM. This is why we also use the predicted hate-label ratio (PHLR), which represents the proportion of all samples labeled as hate. It is calculated as the ratio of the total number of predicted hate labels to the total generated labels, excluding hallucinations and empty outputs.

\textbf{Rectified Scores for Hallucination.} As the aim is to know if the post is hateful, for our use-case, any output not in the form of \textit{`yes/no'}, \textit{`hate/non-hate'}, or \textit{`hateful/non-hateful'} can be considered as a `hallucinated' label falling outside the range of the expected answers. We specify the prompt suffix `answer in one word only.' We also perform a manual evaluation to access the unique outputs per setup, and where the output could be salvaged, they were updated. In line with the existing literature \cite{lee2023crehate}, after all filters, the outputs that still did not qualify as acceptable are discarded. Recording the number of hallucinations (discarded outputs), we introduce `rectified F1/IAA scores', $score_{rectified} = (1-\frac{h}{t})\times score$, where $h$ and $t$ are the hallucinated and total samples, respectively. Any mention of F1 and IAA in our study means `rectified weighted F1' and `rectified IAA.' 

\textbf{Models.} We begin with \texttt{FlanT5} \cite{chung2022scaling}, \mis\ \cite{jiang2023mistral}, \zep\ \cite{tunstall2023zephyr}, \llama\ \cite{touvron2023llama} and InstructGPT \cite{NEURIPS2022_b1efde53} variant \gpt\ (hereby referred to as \cgpt). Performance metrics on all shortlisted LLMs (with the total number of parameters in that model) are captured in Table \ref{tab:base_prompt} for the $p_{base}$. From Table \ref{tab:base_prompt}, we observe that even with $p_{base}$, most models generate noisy and ill-formatted outputs. To reduce the influence of noisy results, we select only \flanxxl\ and \cgpt\ to analyze more intricate setups in our RQs. Both these models portray negligible hallucinations and produce relatively higher F1 and IAA. To test \flanxxl, we employ all the samples. For \cgpt, due to API rate limits, we test random $500$ stratified samples per dataset.

We also examine language-specific LLMs to corroborate the role of zero-shot prompting setup of general-purpose LLM (\cgpt) with cultural context vs language-specific models. The results (Table \ref{tab:lang_llm_try}) highlight that not just demographic-based training corpus but the manner of training and size of models impact multilingualism. 

\textbf{Prompts.} Tables \ref{tab:prompt} and \ref{tab:PromptTrans} provide the list of verbalized prompts under English and multilingual settings, respectively. For the latter, after employing Google Translate, the sample prompts were verified by two linguistic experts.

\textbf{Note on Hate Speech Definition.} Existing literature has observed that providing explicit definitions to LLMs may not lead to performance gain \cite{10459901,roy-etal-2023-probing}. We, thus, conduct our experiments building on the latent notions of hate speech in LLMs. For a sanity check, we prompt \flanxxl\ and \cgpt\ with \textit{`Define: Hate Speech'} and obtain following responses: 
\begin{itemize}[noitemsep,nolistsep,topsep=0pt,leftmargin=2em]
    \item \flanxxl: \textit{``Hate Speech is a form of speech intended to incite hatred or violence against a group of people.''}
    \item \cgpt: \textit{``Hate speech refers to any form of communication, whether spoken, written, or otherwise expressed, that promotes or encourages hatred, hostility, discrimination, or violence against individuals or groups based on attributes such as race, ethnicity, religion, nationality, gender, sexual orientation, disability, or other characteristics. It often involves derogatory language, stereotypes, or dehumanizing rhetoric intended to intimidate, offend, or incite harm against the targeted individuals or communities. Hate speech is typically viewed as a violation of ethical principles and may also be prohibited by law in many jurisdictions due to its potential to inflame tensions, provoke violence, and undermine social cohesion.''}
\end{itemize}

\textbf{Hardware and Software(s).} All experiments are conducted on an NVIDIA-A100 (80GB) GPU. Model weights are loaded from the HuggingFace pretrained models using float16 precision. For \cgpt, OpenAI's API was utilized. All models are employed in a zero-shot setting. The temperature variation experiment takes a total of 100 hours. The rest of the experiments collectively take $\approx$ 4-5 hours of GPU time. 

\begin{table}[!t]
\resizebox{\columnwidth}{!}{
    \begin{tabular}{llcccc}
    \hline
    \textbf{Dataset (Lang)} & \textbf{Model (\# Params)}  & \textbf{\# Samples} & \textbf{\# Hal} & \textbf{F1} & \textbf{IAA} \\ \hline
    \multirow{2}{*}{\mlma\ (Ar)} & \arabicllm\ (7B) & 500 & 378 & 0.027 & 0.0 \\ 
     & \cgpt\ (>150B) & 500 & 222 & 0.359 & 0.140 \\ \hline 
    \multirow{2}{*}{\hasoc\ (Hi)} & \hindillm\ (7B) & 500 & 477 & 0.046 & 0.0 \\
    & \cgpt\ (>150B) & 500 & 131 &  0.257  & 0.018 \\
    \hline
    \end{tabular}}
    \caption{Performance of LLMs when prompted with $p_{base} = $ \textit{`<POST>. Is the given statement hateful?'} in the respective language. We report the number of samples in the dataset used for prompting (\#Samples), the number of hallucinated outputs (\#Hal), and the rectified weighted-F1/inter-annotator agreement (IAA).}
    \label{tab:lang_llm_try}
    \vspace{-3mm}
\end{table}

\section{Do LLMs Pick on Geographical Cues?}
\label{sec:rq1}

\begin{figure*}[t]
\captionsetup[subfigure]{justification=centering}
\subfloat[\flanxxl]{\includegraphics[width=0.32\textwidth]{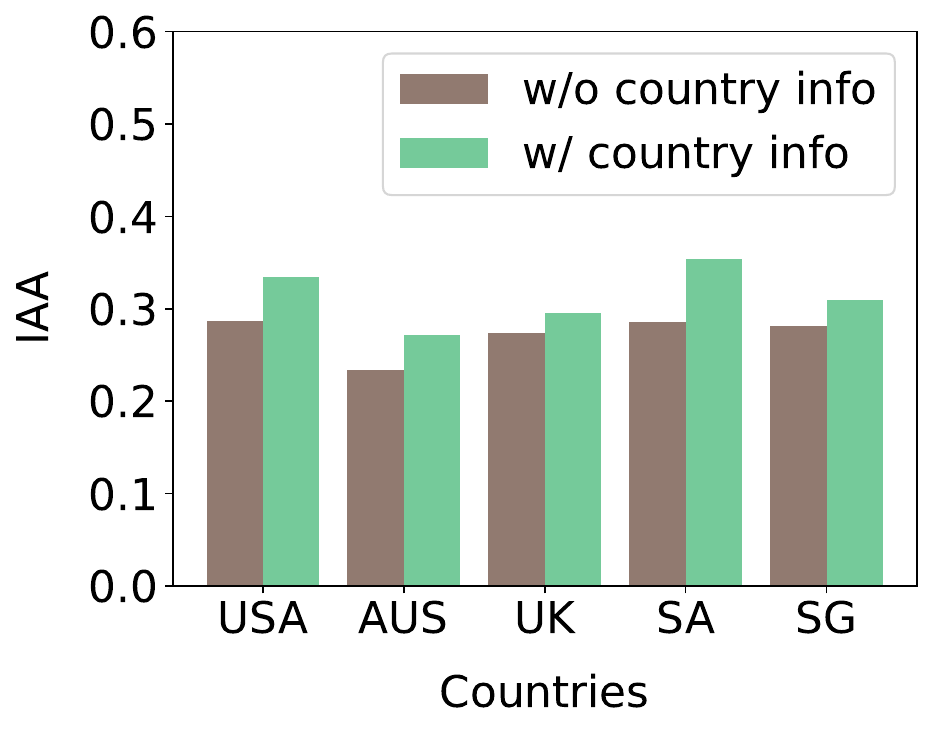}}\hspace{1mm}
\subfloat[\cgpt]{\includegraphics[width=0.32\textwidth]{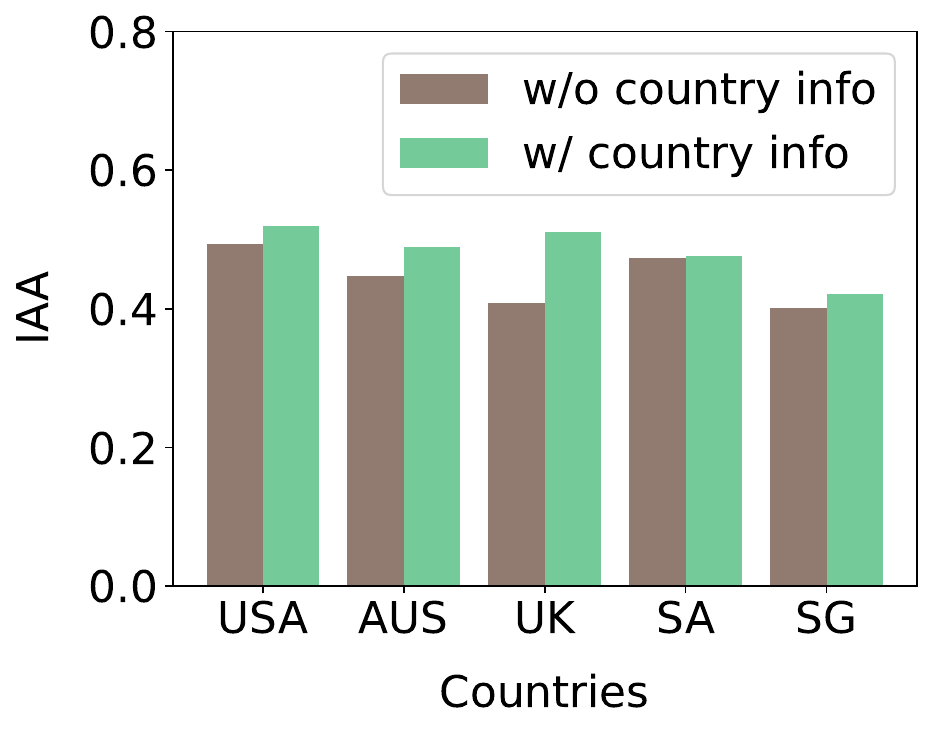}}\hspace{1mm}
\subfloat[\cgpt]{\includegraphics[width=0.32\textwidth]{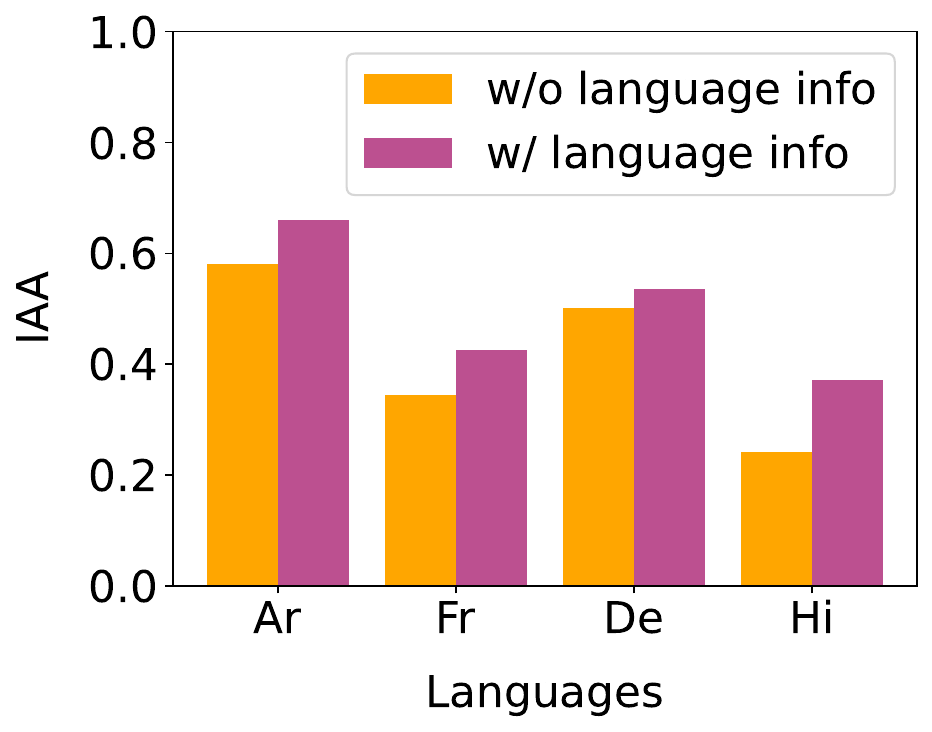}}\hspace{1mm}
\caption{[RQ1] (a-b) The IAA w.r.t human annotation for each country \flanxxl\ and \cgpt, respectively, for English posts. (c) Captures each language's IAA w.r.t human labels via \cgpt~with posts in the language and prompts in English. Here, without (w/o) is $p_{base}$ and with (w/) is $p_{con}$/$p_{lang}$ + $p_{base}$.}
\label{fig:rq_1}
\vspace{-3mm}
\end{figure*}

\textbf{Background.} 
Humans from different countries are prone to variability when flagging the same post as `hateful' or `non-hateful'\cite{lee2023crehate}. Given that AI-assisted content moderation systems often have access to geolocation and language markers along with a post, we investigate whether \textit{`akin to humans, do geographical cues influence LLM's predictions for an incoming hateful post.'}

\textbf{Setup.} Inspired by \citet{lee2023crehate}, we reproduce and extend their analysis on the language tag and multilingual datasets. As described in Section \ref{sec:exp}, we modify their prompt to suit our format. We compare the change in IAA when considering the human annotator of the respective country vs. $p_{base}$ against the human annotator vs. $p_{con} + p_{base}$ or $p_{lang} + p_{base}$. Here, $p_{con}= $ \textit{`The following statement was made in <country>.'}, where $country=$ \textit{\{United States (USA), Australia (AUS), United Kingdom (UK), South Africa (SA), Singapore (SG)\}}. Meanwhile, $p_{lang} = $ \textit{`The following statement was made in <lang> language.'}, where $lang=$\textit{\{Arabic (Ar), French (Fr), German (De), Hindi (Hi)\}}. For analyzing $p_{con}$, we employ \crehate\ for both \flanxxl\ and \cgpt. For $p_{lang}$, we employ the \hasoc\ (German \& Hindi) and \mlma\ (Arabic \& French) on \cgpt's English and language-specific prompting. 

\textbf{Insights.} We discuss the results in two broad 
settings: for country and language tags.

\textbf{Country.} From Figure \ref{fig:rq_1} (b) with $p_{con}$ + $p_{base}$, \cgpt\ exhibits disparity in lack of context about social constructs of the Global South (SA, SG) compared to higher alignment with so-called Western nations (USA, UK, AUS) \cite{zhou-etal-2022-richer, li-etal-2022-herb, lee2023crehate}. Surprisingly, for \flanxxl\ (Figure \ref{fig:rq_1} (a)), we observe an improvement in all countries. It is further corroborated by significance testing on \flanxxl, where we observe (Appendix \ref{app:ptest}) that adding the country cue leads to a notable change in output. One hypothesis for the difference in $p_{con}$ results can be the nature of training the LLM. Due to only instruction tuning \flanxxl\ develops country-specific bias implicitly from the training data; however, for \cgpt\, the implicit bias is augmented with explicit human feedback. It surely calls into question how LLM pretraining mechanisms impact the subjective (non-GLEU) downstream tasks such as hate speech detection \cite{roy-etal-2023-probing}.

\textbf{Language.} From Figure \ref{fig:rq_1} (c), we conjecture that $p_{lang}$ nudges \cgpt\ to language-specific subspaces leading to a visible higher IAA with ground truth labels when we add $p_{lang}$. As we infer in Appendix \ref{app:rq1_multi}, the delta increase in performance in $p_{lang}$ + $p_{base}$ vs. $p_{base}$ follows the same order of magnitude irrespective of whether the prompts are in English or the respective language.     

\textbf{Takeaways.} Firstly, our observations emphasize the fact that both corpus and paradigm in training play a role in the geographical sensitivity of the LLMs. Judging why some cues and LLMs garner more improvement than others is challenging to access from prompting alone. This calls for more transparency in the LLM training to decode the geographical bias during training epochs. Secondly, as an actional insight, the findings encourage incorporating geographical cues in the zero-shot prompt to ensure higher human-LLM alignment. Interestingly, even if the prompt is in the respective language, explicitly nudging helps.

\section{Can LLMs Mimic Annotator Persona?}
\label{sec:rq2}
\textbf{Background.} Socio-cultural experiences of humans color their outlook about hate and cause variation in annotations \cite{sap-etal-2019-risk,orlikowski-etal-2023-ecological}. However, without direct signals about mental state, we rely on demographic attributes as markers of human conditioning \cite{10.1145/3308558.3320096,sap-etal-2022-annotators} a.k.a \emph{personas}. Meanwhile, LLMs only possess statistical socio-cultural experience. When employing LLMs for crowdsourcing, we need to assess \textit{`Do LLM's emulation of demographics at a fine-grained level leads to variations in annotating hate?'}. We further hypothesize that differences in a vulnerable group's projection) can lead to variability in the hate perceptiveness of LLMs. We consider a vulnerable group as those who have historically been mistreated based on identity \cite{10.1145/3580305.3599896}. 

\begin{table*}[t]
\centering
\resizebox{0.95\textwidth}{!}{
\begin{tabular}{llccccccccccccccccc}
\hline
\multirow{3}{*}{\textbf{\begin{tabular}[c]{@{}l@{}}Annotator \\ demographics\end{tabular}}} & \multirow{3}{*}{\textbf{Sub-classes}} & \multicolumn{8}{c}{\textbf{Flan-T5-XXL}} &  & \multicolumn{8}{c}{\textbf{GPT-3.5}} \\ \cline{3-10} \cline{12-19} 
 &  & \multicolumn{2}{c}{$p_{trait}^H$} &  & \multicolumn{2}{c}{$p_{trait}^N$} &  & \multicolumn{2}{c}{$p_{trait}^A$} &  & \multicolumn{2}{c}{$p_{trait}^H$} &  & \multicolumn{2}{c}{$p_{trait}^N$} &  & \multicolumn{2}{c}{$p_{trait}^A$} \\ \cline{3-4} \cline{6-7} \cline{9-10} \cline{12-13} \cline{15-16} \cline{18-19}
 &  & IAA & PHLR &  & IAA & PHLR &  & IAA & PHLR &  & IAA & PHLR &  & IAA & PHLR &  & IAA & PHLR \\ \hline
\multirow{3}{*}{Gender} & Male & 0.42 & 0.53 &  & 0.00 & 0.00 &  & 0.31 & 0.29 &  & 0.40 & 0.70 &  & 0.55 & 0.44 &  & 0.57 & 0.46 \\
 & Female & 0.42 & 0.53 &  & 0.00 & 0.00 &  & 0.33 & 0.31 &  & 0.39 & 0.72 &  & 0.46 & 0.35 &  & 0.52 & 0.54 \\
 & Non-binary & 0.42 & 0.42 &  & 0.01 & 0.01 &  & 0.32 & 0.29 &  & 0.31 & 0.77 &  & 0.45 & 0.38 &  & 0.53 & 0.58 \\ \hline
\multirow{5}{*}{Ethnicity} & Asian & 0.46 & 0.56 &  & 0.03 & 0.02 &  & 0.33 & 0.23 &  & 0.37 & 0.75 &  & 0.55 & 0.51 &  & 0.51 & 0.59 \\
 & Black & 0.43 & 0.61 &  & 0.03 & 0.01 &  & 0.33 & 0.23 &  & 0.37 & 0.74 &  & 0.54 & 0.51 &  & 0.50 & 0.64 \\
 & Hispanic & 0.45 & 0.56 &  & 0.03 & 0.01 &  & 0.36 & 0.24 &  & 0.39 & 0.71 &  & 0.56 & 0.49 &  & 0.51 & 0.62 \\
 & Middle Eastern & 0.46 & 0.52 &  & 0.03 & 0.01 &  & 0.29 & 0.19 &  & 0.40 & 0.70 &  & 0.54 & 0.54 &  & 0.49 & 0.64 \\
 & White & 0.46 & 0.54 &  & 0.03 & 0.01 &  & 0.36 & 0.24 &  & 0.40 & 0.69 &  & 0.51 & 0.57 &  & 0.52 & 0.56 \\ \hline
\multirow{3}{*}{\begin{tabular}[c]{@{}l@{}}Political\\ orientation\end{tabular}} & Liberal & 0.42 & 0.52 &  & 0.01 & 0.01 &  & 0.32 & 0.29 &  & 0.49 & 0.63 &  & 0.54 & 0.53 &  & 0.61 & 0.53 \\
 & Indepedent & 0.43 & 0.58 &  & 0.00 & 0.00 &  & 0.32 & 0.29 &  & 0.46 & 0.65 &  & 0.58 & 0.48 &  & 0.54 & 0.50 \\
 & Conservative & 0.44 & 0.53 &  & 0.00 & 0.00 &  & 0.31 & 0.26 &  & 0.49 & 0.63 &  & 0.53 & 0.56 &  & 0.48 & 0.47 \\ \hline
\multirow{6}{*}{Religion} & Christian & 0.39 & 0.53 &  & 0.03 & 0.03 &  & 0.33 & 0.31 &  & 0.44 & 0.68 &  & 0.53 & 0.58 &  & 0.53 & 0.48 \\
 & Buddhism & 0.41 & 0.54 &  & 0.03 & 0.03 &  & 0.32 & 0.30 &  & 0.45 & 0.66 &  & 0.52 & 0.59 &  & 0.51 & 0.45 \\
 & Islam & 0.43 & 0.49 &  & 0.02 & 0.01 &  & 0.37 & 0.26 &  & 0.47 & 0.65 &  & 0.52 & 0.54 &  & 0.52 & 0.53 \\
 & Judaism & 0.45 & 0.48 &  & 0.02 & 0.01 &  & 0.30 & 0.22 &  & 0.46 & 0.68 &  & 0.50 & 0.60 &  & 0.52 & 0.55 \\
 & Hinduism & 0.42 & 0.49 &  & 0.04 & 0.03 &  & 0.32 & 0.24 &  & 0.48 & 0.65 &  & 0.52 & 0.53 &  & 0.51 & 0.49 \\
 & Irreligious & 0.40 & 0.39 &  & 0.02 & 0.02 &  & 0.30 & 0.25 &  & 0.44 & 0.68 &  & 0.54 & 0.57 &  & 0.54 & 0.43 \\ \hline
\multirow{6}{*}{\begin{tabular}[c]{@{}l@{}}Education\\ level\end{tabular}} & \textless High school & 0.42 & 0.46 &  & 0.01 & 0.01 &  & 0.31 & 0.30 &  & 0.52 & 0.60 &  & 0.54 & 0.51 &  & 0.50 & 0.49 \\
 & High school & 0.41 & 0.52 &  & 0.00 & 0.01 &  & 0.29 & 0.29 &  & 0.45 & 0.65 &  & 0.56 & 0.52 &  & 0.53 & 0.50 \\
 & College & 0.42 & 0.52 &  & 0.00 & 0.01 &  & 0.29 & 0.29 &  & 0.44 & 0.67 &  & 0.53 & 0.44 &  & 0.52 & 0.46 \\
 & Bachelor & 0.42 & 0.52 &  & 0.00 & 0.00 &  & 0.28 & 0.29 &  & 0.43 & 0.68 &  & 0.56 & 0.47 &  & 0.50 & 0.44 \\
 & Master's & 0.42 & 0.52 &  & 0.00 & 0.00 &  & 0.28 & 0.29 &  & 0.41 & 0.69 &  & 0.54 & 0.45 &  & 0.48 & 0.46 \\
 & PhD & 0.42 & 0.51 &  & 0.00 & 0.01 &  & 0.27 & 0.28 &  & 0.47 & 0.65 &  & 0.53 & 0.42 &  & 0.51 & 0.49 \\ \hline
\end{tabular}}
\caption{[RQ2] The inter-annotator agreement (IAA) and predicted hate label ratio (PHLR) w.r.t majority voted gold label in \crehate, for \flanxxl\ and \cgpt. The demographic attributes ($D$) are compared under the hateful $p_{trait}^H$, non-hateful $p_{trait}^N$, and assumed persona $p_{trait}^A$ settings. $\forall D_* \in D$, we combine the $p_{trait}^*$ + $p_{base}$.   
}
\label{tab:rq2}
\end{table*}

\begin{figure*}[!t]
\centering
\captionsetup[subfigure]{justification=centering}
\subfloat[\cgpt\ (H)]{\includegraphics[width=0.25\textwidth]{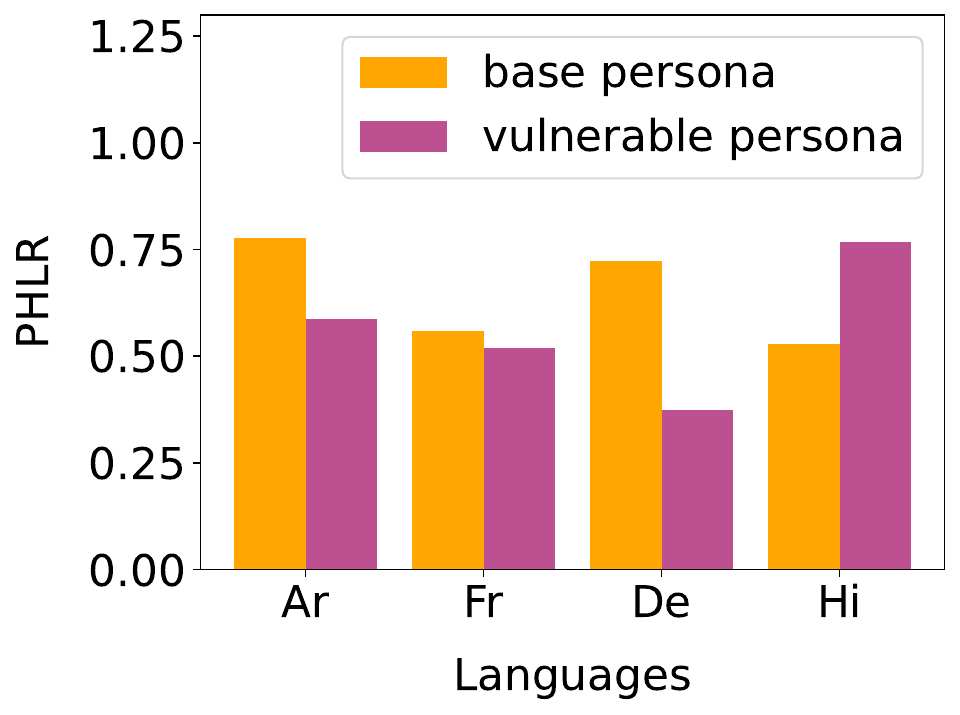}}\hfil
\subfloat[\cgpt\ (N)]{\includegraphics[width=0.245\textwidth]{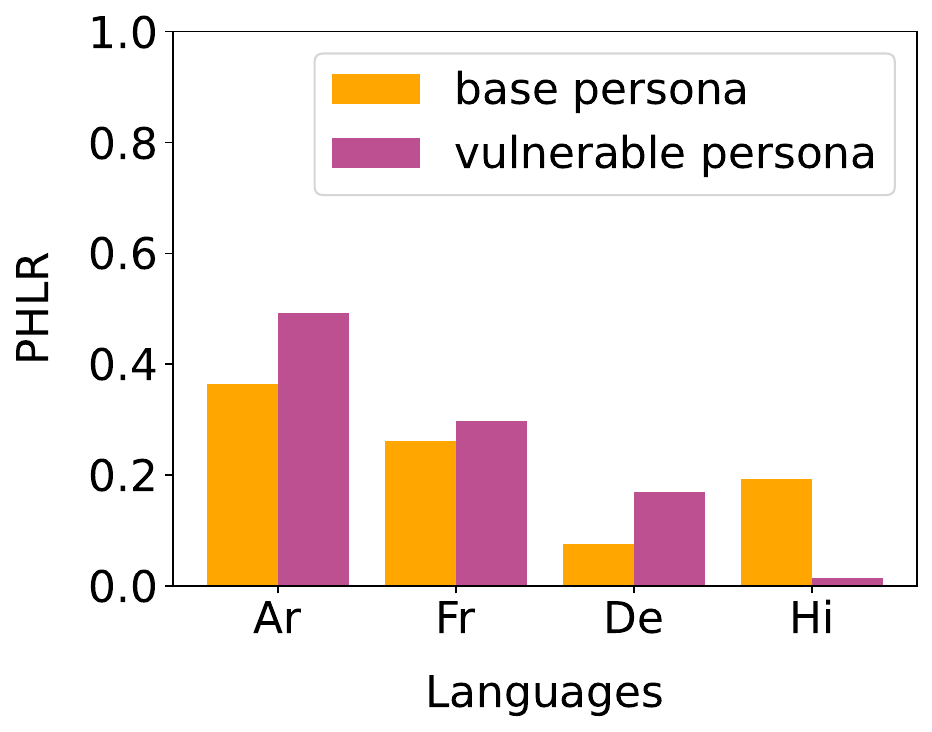}}
\subfloat[\cgpt\ (H)]{\includegraphics[width=0.25\textwidth]{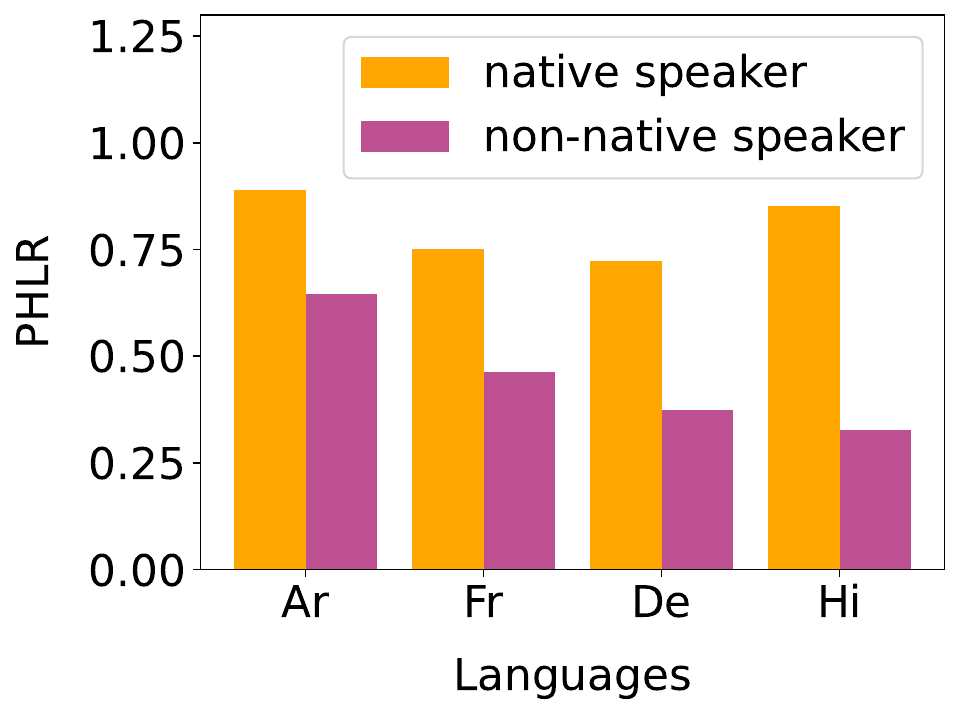}}\hfil
\subfloat[\cgpt\ (N)]{\includegraphics[width=0.24\textwidth]{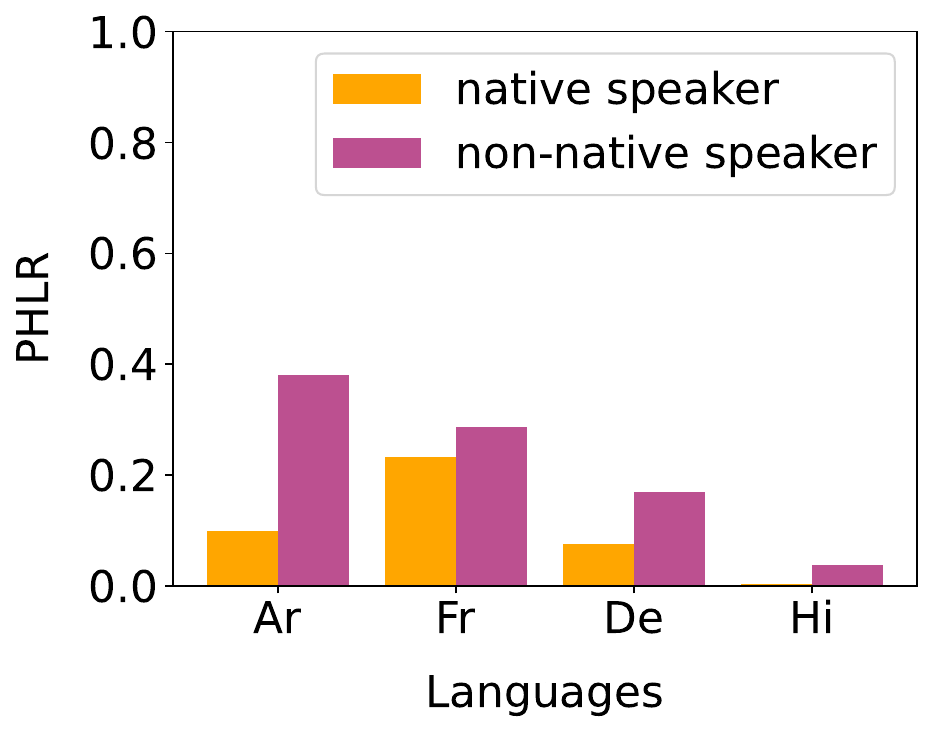}}
\vspace{-3mm}
\caption{[RQ2] Predicted hate label ratio (PHLR) from \cgpt\ comparing $p_{trait}^{L_*}$ + $p_{base}$ for Arabic, French, German, and Hindi. (a) and (b) capture the base vs. vulnerable persona for hate/non-hate queries, respectively. (c) and (d) capture the native vs. non-native speaker persona for hate/non-hate queries, respectively.}
\label{fig:rq_2}
\vspace{-3mm}
\end{figure*}

\textbf{Setup.} Borrowing the annotator demographic list from \crehate, we define our persona attributes in terms of $D$=\textit{\{Gender ($D_g$), Ethnicity ($D_e$), Political Orientation ($D_p$), Religion ($D_r$), Education Level ($D_q$)\}}. Each demographic in $D_* \in D$ has further sub-classes as outlined in Table \ref{tab:rq2}. For each $D_*$, we augment the prefix $p_{trait} = $\textit{`A person who is <D*>, annotated the following statement as <H/N>.'} Operated via the <H/N> tag, we run two persona variants, calling the annotated statement either hateful $p_{trait}^H$ or non-hateful $p_{trait}^N$. We also examine a third variant $p_{trait}^A$ with the prompt $p_{trait}^A= $\textit{`Suppose you are a person who is <D> and given the following statement to annotate.'} The $3$ personas prompts ($p_{trait}^*$ + $p_{base}$) are examined via the \crehate\ dataset on \flanxxl\ and \cgpt.

Intrigued by the success of the language tag in RQ1 (Section \ref{sec:rq1}), we deep dive into persona traits that closely represent each language's demographic. Here, $\forall l \in lang=$ \textit{\{Arabic, French, German, Hindi\}}, we introduce prompts $p_{trait}^{L_*} = $ \textit{`The following statement is in <l> language, and a <I> annotated this as <H/N>.'} In the prompt, $I$ enlists a base/majoritarian persona vs. a vulnerable/minority persona of the respective geography examined on \cgpt. Steps for obtaining the following personas are provided in Appendix \ref{app:rq2_multi}.
\begin{enumerate}[noitemsep,nolistsep,topsep=0pt,leftmargin=2em]
    \item For Arabic, 
    \resizebox{4cm}{!}{$p_{trait}^{L_{Ar}} \in  \{$\textit{Muslim/Non-mulsim}$\}$}
    \item For French, 
    \resizebox{5cm}{!}{$p_{trait}^{L_{Fr}} \in \{$\textit{French/Mediterranean descent}$\}$}
    \item For German, 
    \resizebox{4.8cm}{!}{$p_{trait}^{L_{De}} \in \{$\textit{Native/Non-native German speaker}$\}$}
    \item For Hindi, 
    \resizebox{4cm}{!}{$p_{trait}^{L_{Hi}} \in  \{$\textit{Upper/Lower caste}$\}$}
\end{enumerate}
We examine the hateful ($H$) and non-hateful ($N$) queries here as well, both with English and multilingual prompts. As none of the datasets provide demographic-specific labels, we rely on IAA among predicted and the majority-voted gold labels for our assessment. For $p_{trait}^{H/N}$, we provide in the prompt if the persona identifies a statement as hate/non-hate speech. Our objective here is to assess the role of these traits in persuading LLMs to increase/decrease the number of hate labels in their responses. We thus utilize the Predicted Hate Label Ratio (PHLR) metric in addition to IAA.

\textbf{Insights.} We discuss the results broadly for English and multilingual datasets.

\textbf{English:} Table \ref{tab:rq2} corroborates that nudging the model to assume a persona ($p_{trait}^A$) is different from presenting the LLMs with a persona ($p_{trait}^{H/N}$). Further, it is evident from the predicted hate-label ratio (PHLR) that LLMs are sensitive towards some demographic subclasses more than others. Significance testing corroborates the same (Appendix \ref{app:ptest}).

Under Gender demographic for \cgpt, the percentage of hate labels is higher for `Non-binary' than `Males' (for $p_{trait}^H$). It aligns with the former being a more susceptible subclass of gender in the real world. Meanwhile, for \flanxxl, the presence of the non-hate tag $p_{trait}^N$ dominates the demographic information in the context. However, the opposite is not valid for $p_{trait}^H$. Despite the ground labels associated with samples being balanced across classes, $p_{trait}^N$ for \flanxxl\ still predicts the majority of the labels as `non-hate.'

\begin{figure*}[!t]
\captionsetup[subfigure]{justification=centering}
\subfloat[\flanxxl~(H)]{\includegraphics[width=0.25\textwidth]{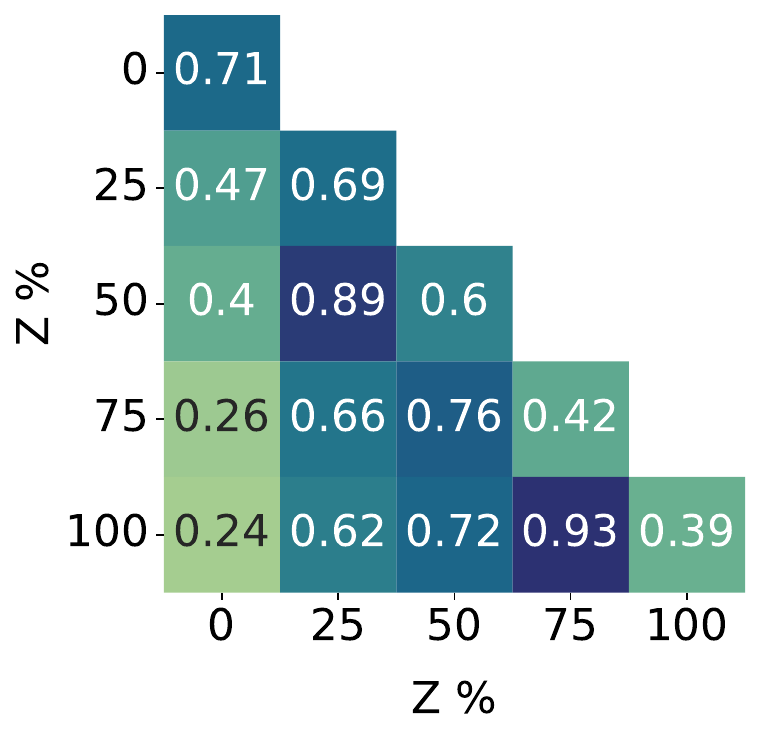}}
\subfloat[\cgpt~(H)]{\includegraphics[width=0.25\textwidth]{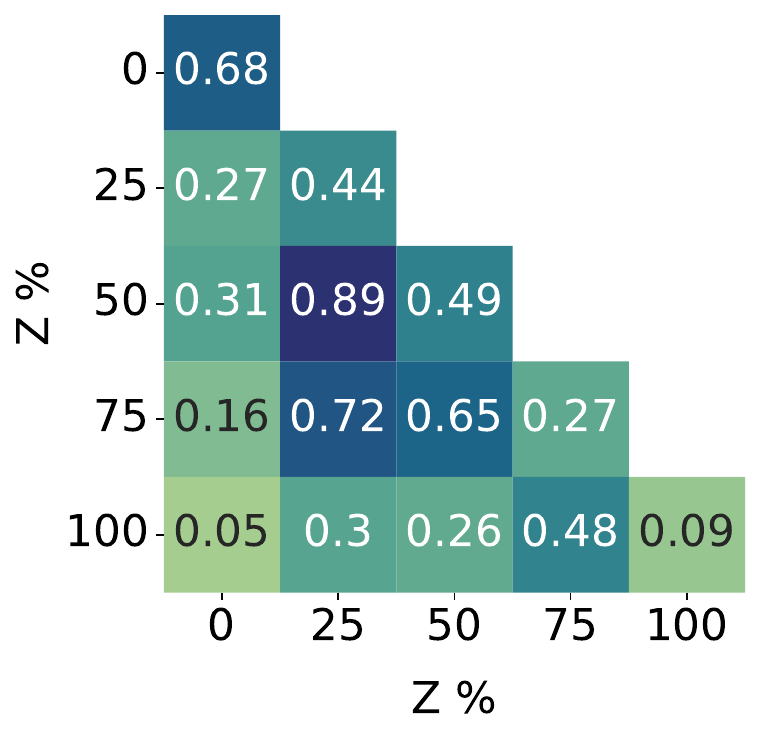}}
\subfloat[\flanxxl~(N)]{\includegraphics[width=0.25\textwidth]{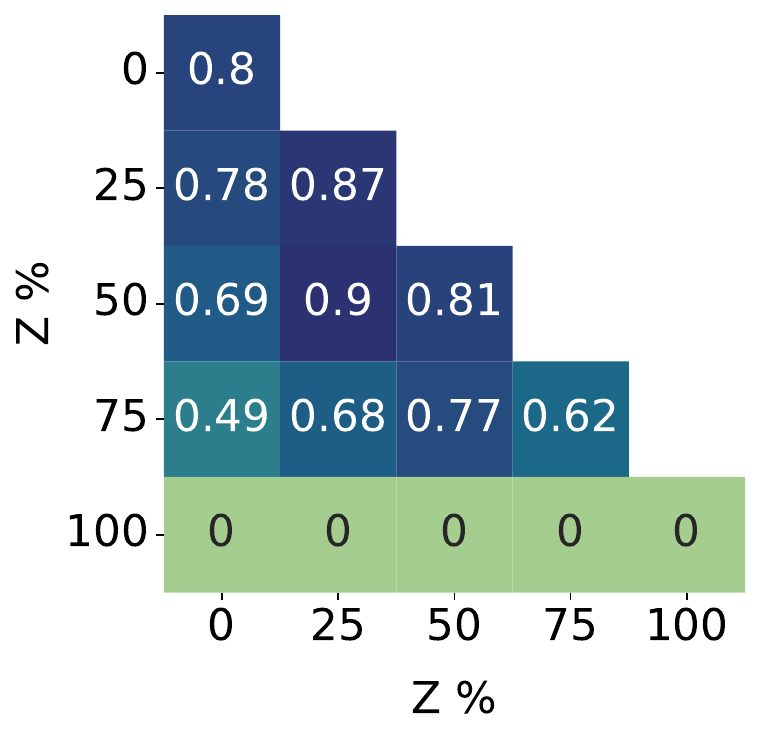}}
\subfloat[\cgpt~(N)]{\includegraphics[width=0.25\textwidth]{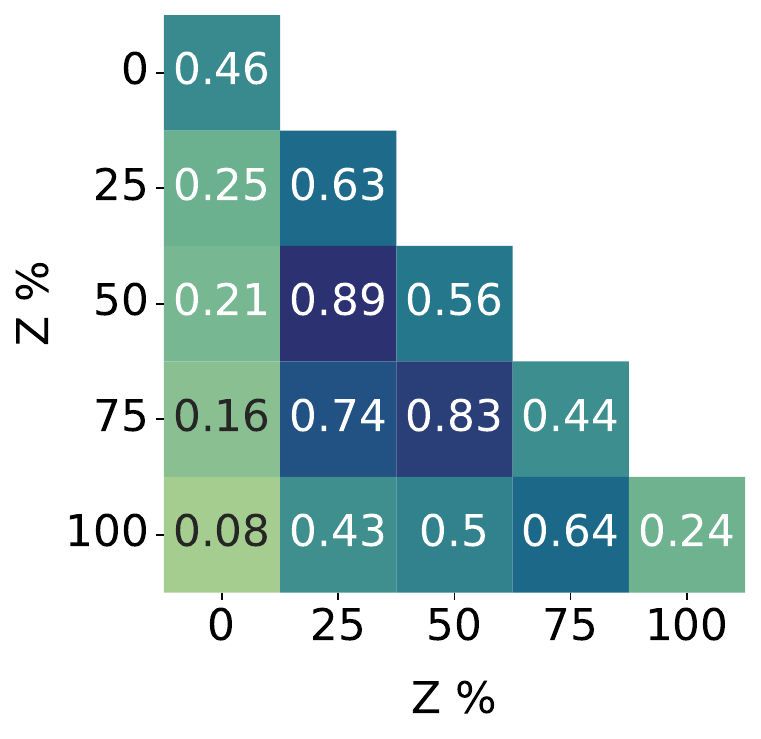}}
\caption{[RQ3] For the $p_{vote}^H$ : (a) and (b) capture the IAA  among various hateful voting percentages, i.e., $z \%$ for \flanxxl\ and \cgpt, respectively. For $p_{vote}^N$: (c), (d) function analogously to (a), (b), respectively.} Note: diagonals in heatmaps represent IAA between $p_{base}$ and $p_{vote}$ + $p_{base}$. Meanwhile, when x and y are different, it represents IAA b/w ($P_{vote}=x + P_{base}$) and ($P_{vote}=y + P_{base}$).
\label{fig:rq_3}
\vspace{-3mm}
\end{figure*}

\textbf{Multilingual.} From Figure \ref{fig:rq_2} (a-b), we hypothesize that explicitly known vulnerabilities like Islamophobia (Arabic + Muslim) and Casteism (Hindi + Lower caste) are better captured by the LLM than ethnicity, leading to higher sensitivity of these pairs towards hate when prompted in English. While the patterns persist under multilingual prompting for the rest of the languages, its lack of variation in $p_{trait}^H$ Hindi is puzzling (Table \ref{tab:rq2_multi_1} in Appendix \ref{app:rq2_multi}).

As the highest gap in $p_{trait}^*$ English is observed for German, we deep dive into this prompt (native speaker vs non-native speaker) and repeat this for all languages under English and multilingual prompting. From Figure \ref{fig:rq_2} (c-d) and  Table \ref{tab:rq2_multi_2} in Appendix \ref{app:rq2_multi}, one can conjecture that when a native-speaking persona considers a post as hateful, the model may be contextualizing the higher acuity of the speaker to understand the geographical context.

\textbf{Takeaways.} Given the sensitivity of LLMs toward the combination of persona + label, we urge the practitioners not mindlessly to rely on LLM as a substitute for demographic attributes. The persistence of results in English and multilingualism is an advantage for the researchers working on multilingual datasets. It can allow faster assessment of the LLMs without requiring separate prompt engineering for both setups.

\section{Are LLMs Sensitive to Anchoring Bias?}
\label{sec:rq3}
\textbf{Background.} Anchoring bias occurs when humans rely too heavily on the anchor information (relevant or not) to influence their decision-making. While some anchors like language and geolocation, as established in RQ1 (Section \ref{sec:rq1}), are helpful for LLM-based content moderation, the influence of numerical cues under the zero-shot setting is unknown. Numerical features/cues can be defined as aggregated real-world or simulated values representing countable metadata associated with the posts. It can range from the total/average views, likes, and comments a post/user receives to the number of people who have in the past reported/flagged the post/user as hateful. The former set of metadata is publicly available and has been successfully employed in hate speech detection to improve finetuned accuracy \cite{10.1145/3292522.3326028, 10.1145/3580305.3599896}. Meanwhile, the influence of previously flagged counts is not known, as such metrics are not available to the public. Here, using voting statistics as a proxy for the crowd's opinion about the post, we are motivated to examine whether `\textit{made-up voting percentages in the prompt lead to manipulation in LLM's output?}'

\textbf{Setup.} The base prompt $p_{base}$, is prefixed with $p_{vote}^{H/N}= $ \textit{`The following statement was labeled <H/N> by <z> annotators.'}, where $z\in\{0\%, 25\%, 50\%, 75\%, 100\%\}$. The two variants $p_{vote}^{H/N}$ capture the hateful (H) or non-hateful voting (N) label. We represent $z$ in the percentage to give a relative sense of majority voting. At $z=50\%$ saying, 50\%, annotators consider the post is hateful, and 50\% say it is non-hateful is not the same as $p_{base}$. We conduct this experiment with \hatexp\ on \flanxxl\ and \cgpt\ with $p_{vote}$ + $p_{base}$. The $z$\% alludes to annotators in general and not a specific persona. Further, it should be noted that these percentages are not available as a part of \hatexp, and added by us to introduce the ``anchoring" information.  

\textbf{Insights.} In line with existing evidence \cite{pmlr-v139-zhao21c} of majority labeling bias in few-shot learning, we also establish that LLMs are prone to labeling bias even under zero-shot settings if the context mentions voting percentages. From Figures \ref{fig:rq_3} (a-b) regarding $p_{vote}^H$, it is evident that alignment in hate labeling is more consistent when the percentages lie closer to each other and decrease as one moves away. Succinctly, $IAA_{ij}>IAA_{ik}$ if $|z_j-z_i| < |z_k - z_i|$. A similar pattern is observed for $p_{vote}^N$ (Figures \ref{fig:rq_3} (c-d)). 

\textbf{Takeways.}
Corroborated by significance testing (Appendix \ref{app:ptest}) and controlling for decoding temperature (Appendix \ref{app:rq3}), we discover that while LLMs understand relative percentages, they are prone to emphasize this information over the post's content. As in our RQ, voting serves as a proxy of numerical metadata in the content moderation pipeline; it implies that corruption (accidental or intentional) can lead to instance misclassification. Despite numerical/statistical features being used earlier to enhance the finetuning in hate speech detection \cite{10.1145/3292522.3326028,10.1145/3580305.3599896}, they cannot be directly extrapolated as features in prompting for hate speech annotations.

\section{Discussion}
This section summarizes the implications of the research. We also provide recommendations for LLM-assisted content moderation. Except for the \flanxxl\ non-hate persona, the results across models, datasets, languages, and prompts are aligned. Cognizant of adversarial attacks on prompting, the study balances both feature importance (corroborated by significance testing) and cautions against adversary features. Despite how the perception of hatefulness manifests via  $p_{base}$, we observe that the inclusion of cues nudges the LLMs to explore a contextual definition of hate.

\textbf{Geographical Sensitivity.} How the latent spaces are triggered at the mention of geographical cues \cite{zhou-etal-2022-richer} is intractable from prompting alone. This becomes especially tricky for overlapping signals like `Arabic,' `Muslims,' and `Islamophobia' (Figure \ref{fig:rq_1} (c)). It once again highlights the need for more transparency in the training and cultural alignment of LLMs. As a quick fix, we suggest augmenting the language/country tags for an incoming post being prompted to improve alignment with people representative of a given region/geography. 

\textbf{Demographic Sensitivity.} Our work establishes that the manner of personification of demographic attributes can lead to variation in hate labeling, with assumed persona ($p_{trait}^A$) being closer to the inherent knowledge and biases an LLM possesses. Here, we caution practitioners looking to adopt LLMs for crowdsourcing to experiment with different framing of the personas and identity traits.

\textbf{Numerical Sensitivity.} As evident from our experiments, LLMs do not have a clear way of discovering noise from the informativeness in the prompt. Our advice to practitioners is to refrain from adding any numerical statistics in the prompt unless they are quality-checked and to mask such adversarial expressions written by the authors of the input samples.

\textbf{Multilingual Sensitivity.} Following the same prompt format, we observe similar performance patterns across both English and Multilingual settings (refer Table \ref{app:rq_1_multi} in Appendix \ref{app:rq2_multi}). The results are encouraging, allowing content moderators to work in the native language of the post without extraneous prompt engineering. The degradation in performance from English to Multilingual \cite{jin2023better}, however, calls for more investment in non-English training and evaluation of LLMs.

\textbf{LLMs for Hate Speech Annotations.} It is imperative to point out that hate speech is a human-centric phenomenon steeped in historical and cultural contexts. As such, any computational attempt to flag it can only be assistive. Our experiments over three contextual settings indicate that LLM cannot outright substitute a demographic group in the annotation process (Section \ref{sec:rq2}). Further, if LLMs are prone to anchoring, that is not always beneficial. Then, our study opens up the questions around using few-shot/guideline examples for labeling hate (Sections \ref{sec:rq2} and \ref{sec:rq3}).

\section{Conclusion}
In the current LLM space, both alignment and robustness do not have a singular non-overlapping definition, which makes our study sit at the intersection of robustness in hate speech detection as well as conceptualization of the human-LLM alignment. Over multiple RQs and prompt setups, we explore how well-suited LLMs are for assisting humans at different stages of the content moderation pipeline. Our results on demographic sensitivity, cultural priming, and anchoring bias are evident over multiple annotations on 6 datasets, 5 languages, and 2 LLMs. Our analysis reiterates using LLMs as an assistive system rather than substituting human moderators. We would like to work on more datasets and explore intersectional demographic attributes in the future.  

\newpage
\clearpage
\section{Limitations}
First and foremost, the list of research questions and prompts analyzed, while significant in number, is not exhaustive. Given that the adoption of LLMs in the hate speech moderation pipeline is nascent, the scope for research is more vast than what we can account for in current research. We hope our findings and primary analysis motivate future research. Besides, there is a shortage of open-source instruction-tuned multilingual LLMs, which restricts our multilingual and multicultural analysis to \cgpt. Owing to the lack of datasets with ground truth labels where multiple annotators are considered and demographic-specific annotators are released, performing one-to-one mapped inter-annotator agreement analysis between LLMs and demographic personas is challenging. The current study only utilizes textual content. In the future, we need to conduct assessments for content moderation on other modalities like memes, short videos, etc. It should be noted that geography or language only acts as a proxy for cultural and linguistic recalls and does not represent an absolute assessment for multilingual LLM evaluations. Consequently, when employing geographical or demographic attributes, we need to be cognizant of the fact that the sensitivity of the LLMs can also be a source of their implicit bias.

\section{Ethical Considerations}
This work does not produce any new scientific artifacts regarding datasets or proposed models, and no human evaluators were involved. Our work utilizes LLMs to annotate \cite{doi:10.1073/pnas.2305016120} hate speech, which comes with legal, technical, and ethical challenges. There is an inherent issue of magnifying human behaviors and biases by LLMs. Our findings corroborate the sensitivity of LLMs towards different demographics and identity traits. Hence, their generations/predictions should be taken cautiously. Further, this study is underpinned by gold labels obtained from human annotators, which themselves could be erroneous. The intent of releasing our prompt list is to broaden the examination of biases and fallacies an LLM is prone to. While \flanxxl\ is an opensource LLM, \cgpt\ is not, which raises concerns around reproducibility \cite{10.1038/s42256-023-00783-6}, and is a broader challenge for the NLP community. 

\section*{Acknowledgements}
Sarah Masud acknowledges the support of the Prime Minister Doctoral Fellowship in association with Wipro AI and
Google PhD Fellowship.
Viktor Hangya and Alexander Fraser acknowledge funding from the German Research Foundation (DFG; grant FR 2829/7-1). Tanmoy Chakraborty acknowledges Anusandhan National Research Foundation (CRG/2023/001351) for the financial support. 
The authors thank Ahmed Tarek Sobhi Abdelrahman and  Nicolas Payen for validating the French, German, and Arabic prompts. 

\bibliography{emnlp24}
\clearpage
\newpage
\appendix
\section{RQ1 Multilingual Prompting}
\label{app:rq1_multi} 
We run additional experiments where the cues and query are in the same language as the multilingual post. The comparative results are enlisted in Table \ref{app:rq_1_multi}. Despite the expected loss in performance \cite{jin2023better}, the inclusion of $p_{lang}$ leads to a better human-AI alignment. 

\section{RQ2 Multilingual Persona Prompting}
\label{app:rq2_multi}
\textbf{Personas.} We start with Wikipedia to get a general sense of the demographics of each geography. From there, we narrow down the most prominent demography of the nation. We further use census and news articles and consult social experts about each geography before narrowing down the vulnerable minority based on the hypothesis that minority groups get more hate than the majority.

\textbf{Prompting.} From Tables \ref{tab:rq2_multi_1} and \ref{tab:rq2_multi_2}, we again observe the same pattern as in English, thereby showcasing that the role of context is significant irrespective of the language under consideration. Although we have seen a similar pattern in PHLR, there is a significant degradation in the observed metrics compared to English. 

\section{RQ3 Temperature Probing}
\label{app:rq3}
\begin{figure*}[!h]
\captionsetup[subfigure]{justification=centering}
\centering
\subfloat[\flanxxl\ temperatures (H)]{\includegraphics[width=\columnwidth]{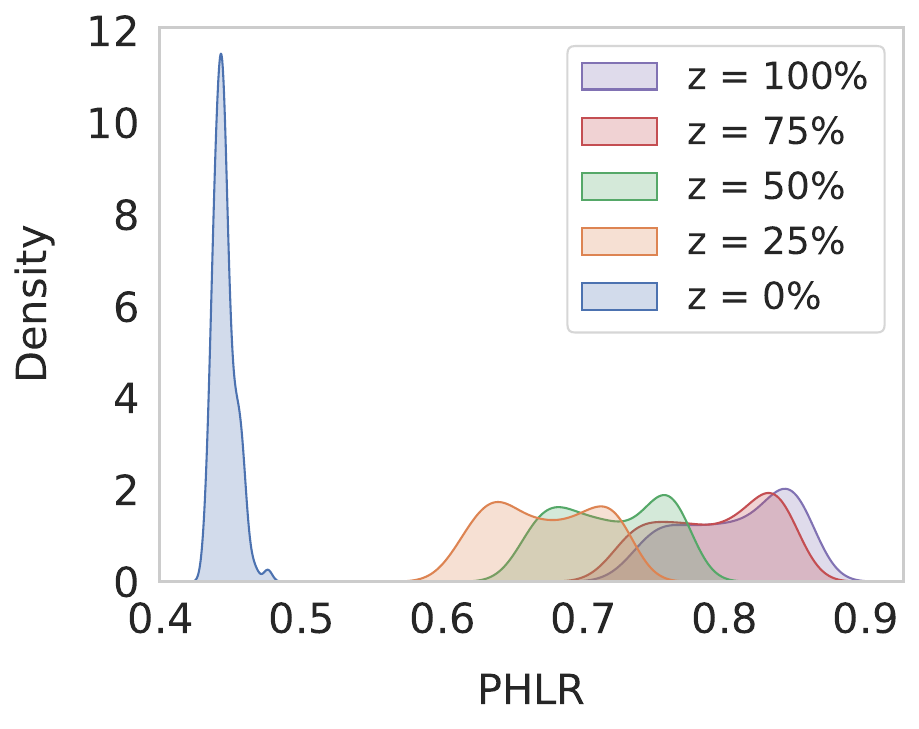}} \hspace{1mm}
\subfloat[\flanxxl\ temperatures (N)]
{\includegraphics[width=\columnwidth]{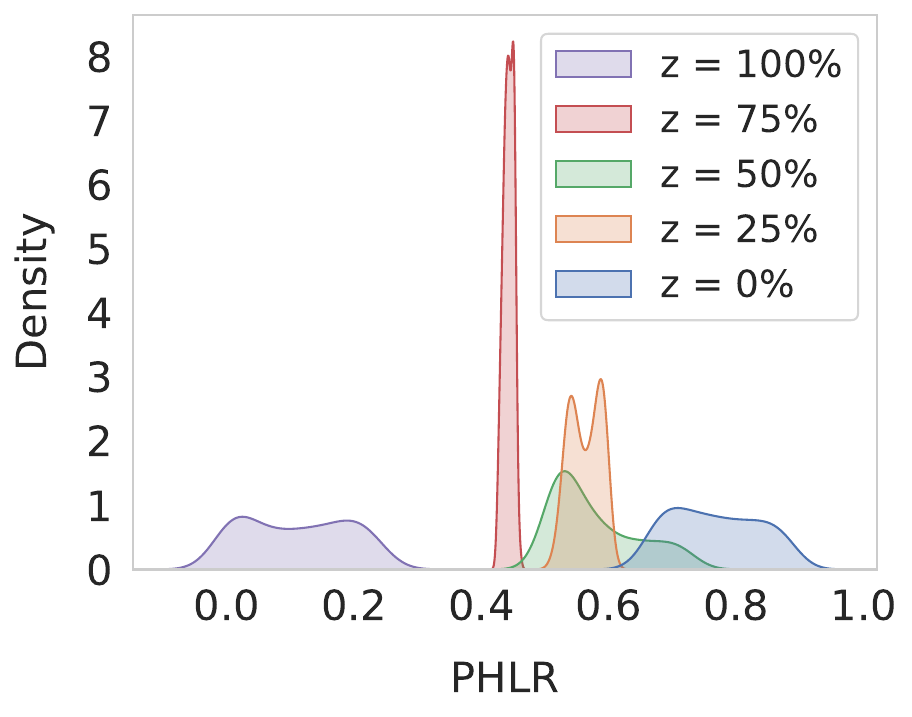}}
\caption{[RQ3] The impact of decoding temperature with varying voting percentages for $p_{context}^H$ on \flanxxl.}
\label{fig:rq_1_temp}
\vspace{-3mm}
\end{figure*}

We simulate multiple annotators under each $z$ value to further corroborate the results. Due to resource constrain, we run this analysis on only \flanxxl. We generate $100$ output for each sample, by uniformly sampling $100$ decoder temperature values ($t\in(0, 2)$). For reference, in $p_{base}$, we obtain a mean (std. dev) percentage of hate label as $0.560 (\pm 0.0169)$. While the spread is low for $p_{base}$, we observe from Figure \ref{fig:rq_1_temp} (a) and (b) that for some $z$, the spread varies. It is an indicator that decoding temperature can lead to variations in the LLM's predictions when prompted on numerical anchors. We observe more spread on average when using $p_{vote}$ as recorded in Table \ref{tab:mean_sd_dist}. For $p_{vote}^H$, with varying temperatures, as the value of $z$ increases, the percentage of hate predicted increases, as evident from the shift towards the right for $z=100\%$ in Figure \ref{fig:rq_1_temp} (a). The reverse trend is observed for $p_{vote}^N$ in Figure \ref{fig:rq_1_temp} (b) where the curve for $z=100\%$ is left shifted, leading to a decrease in the percentage of hate predicted as the majority of non-hateful increases.

\section{Statistical Testing on FlanT5-XXL}
\label{app:ptest}

\textbf{Process.} For RQ1 (Sec. \ref{sec:rq1}) and RQ3 (Sec. \ref{sec:rq3}), we perform the paired t-test and report the effect size. Meanwhile, for RQ2 (Sec. \ref{sec:rq2}), to capture the intra-demography disparity among the subclasses, we use ANOVA. All experiments are run via the SciPy and NumPy libraries.

\textbf{Observations.} For all the RQs, we observe significant disparity caused by context.
\begin{itemize}[noitemsep,nolistsep,topsep=0pt,leftmargin=2em]
    \item \textbf{RQ1:} From Table \ref{tab:rq1_stat}, comparing both adjusted F1 and IAA, we observe that adding the country cue leads to a significant change in the prompted output (as captured by the higher effect size (ES) as well as the p-values). Following the reference Figure \ref{fig:rq_1} (a), South Africa registers the highest impact.
    \item \textbf{RQ2:} From Table \ref{tab:rq2_stat} we observe that different modes of the persona ($p^H$, $p^N$, $p^A$) are impacted by varying sub-classes. Interestingly, the subgroups within Religion show considerable variation for three persona prompts.
    \item \textbf{RQ3:} From Table \ref{tab:rq3_stat}, for both $p^H$ and $p^N$, we observe significant differences (as captured by the higher effect size (ES) as well as the p-values) in performance among the various percentages in the prompt. This further corroborates the anchoring bias in LLMs employed for zero-shot labeling.
\end{itemize}

\begin{table}[!h]
\resizebox{\columnwidth}{!}{
\begin{tabular}{lcclcc}
\hline
\multirow{2}{*}{\textbf{Language}} & \multicolumn{2}{c}{\textbf{\begin{tabular}[c]{@{}c@{}}Prompt in\\ English\end{tabular}}} &  & \multicolumn{2}{c}{\textbf{\begin{tabular}[c]{@{}c@{}}Prompt in \\ same language\end{tabular}}} \\ \cline{2-3} \cline{5-6} 
 & $p_{base}$ & $p_{lang}$ &  & $p_{base}$ & $p_{lang}$ \\ \hline
Arabic & 0.580 &	0.660 &	& 0.140 &	0.305 \\
French & 0.344 & 0.425 &  & 0.272 & 0.356 \\
German & 0.502 & 0.537 &  & 0.412 & 0.423 \\
Hindi & 0.242 & 0.371 &  & 0.018 & 0.031 \\ \hline
\end{tabular}}
\caption{[RQ1] Extension of Figure \ref{fig:rq_1} (c) for IAA comparison of the results without and with language cues for prompts in English and the respective language.}
\label{app:rq_1_multi}
\vspace{-3mm}
\end{table}

\begin{table}[!h]
\centering
\resizebox{\columnwidth}{!}{
\begin{tabular}{llcclcc}
\hline
\multirow{2}{*}{\textbf{Language}} & \multirow{2}{*}{\textbf{Majority or vulnerable}} & \multicolumn{2}{c}{\textbf{\begin{tabular}[c]{@{}c@{}}Prompt in\\ English\end{tabular}}} &  & \multicolumn{2}{c}{\textbf{\begin{tabular}[c]{@{}c@{}}Prompt in\\ same language\end{tabular}}} \\ \cline{3-4} \cline{6-7} 
 &  & $p^H$ & $p^N$ &  & $p^H$ & $p^N$ \\ \hline
\multirow{2}{*}{Arabic} & Muslim & 0.778 & 0.364 &  & 0.992 & 0.737 \\
 & Non-muslim & 0.586 & 0.492 &  & 0.525 & 0.483 \\ \hline
 \multirow{2}{*}{French} & French descent & 0.558 & 0.262 &  & 0.666 & 0.170 \\
 & Mediterranean descent & 0.520 & 0.298 &  & 0.649 & 0.176 \\ \hline
\multirow{2}{*}{German} & Native & 0.724 & 0.076 &  & 0.566 & 0.152 \\
 & Non-native & 0.374 & 0.170 &  & 0.248 & 0.248 \\ \hline
\multirow{2}{*}{Hindi} & Upper caste & 0.528 & 0.192 &  & 0.998 & 0.282 \\
 & Lower caste & 0.768 & 0.014 &  & 0.998 & 0.094 \\ \hline
\end{tabular}}
\caption{[RQ2] PHLR for comparison of vulnerable persona cues. Extension of Figure \ref{fig:rq_2} (a-b).}
\label{tab:rq2_multi_1}
\vspace{-3mm}
\end{table}

\begin{table}[!h]
\resizebox{\columnwidth}{!}{
\begin{tabular}{llcclcc}
\hline
\multirow{2}{*}{\textbf{Language}} & \multirow{2}{*}{\textbf{Speaker}} & \multicolumn{2}{c}{\textbf{\begin{tabular}[c]{@{}c@{}}Prompt in\\ English\end{tabular}}} &  & \multicolumn{2}{c}{\textbf{\begin{tabular}[c]{@{}c@{}}Prompt in\\ same language\end{tabular}}} \\ \cline{3-4} \cline{6-7} 
 &  & $p^H$ & $p^N$ &  & $p^H$ & $p^N$ \\ \hline
\multirow{2}{*}{Arabic} & Native & 0.890 & 0.100 &  & 0.764 & 0.099 \\
 & Non-native & 0.646 & 0.380 &  & 0.567 & 0.901 \\ \hline
 \multirow{2}{*}{French} & Native & 0.752 & 0.232 &  & 0.916 & 0.130 \\
 & Non-native & 0.462 & 0.286 &  & 0.702 & 0.106 \\ \hline
\multirow{2}{*}{German} & Native & 0.724 & 0.076 &  & 0.566 & 0.152 \\
 & Non-native & 0.374 & 0.170 &  & 0.248 & 0.248 \\ \hline
\multirow{2}{*}{Hindi} & Native & 0.852 & 0.004 &  & 1.000 & 0.753 \\
 & Non-native & 0.328 & 0.038 &  & 1.000 & 0.858 \\ \hline
\end{tabular}}
\caption{[RQ2] PHLR for comparison of speaker persona cues. Extension of Figure \ref{fig:rq_2} (c-d).}
\label{tab:rq2_multi_2}
\vspace{-3mm}
\end{table}

\begin{table}[h]
\centering
{
\begin{tabular}{lcc}
\hline
\multirow{2}{*}{\textbf{<z>}} & \multicolumn{2}{l}{\textbf{Distribution parameters ($\mu \pm \sigma$)}} \\ \cline{2-3} 
 & $p_{vote}^H$ & $p_{vote}^N$ \\ \hline
0\% & $0.445 \pm 0.0078$ & $0.77 \pm 0.067$ \\
25\% & $0.67 \pm 0.037$ & $0.56 \pm 0.024$ \\
50\% & $0.72 \pm 0.037$ & $0.57 \pm 0.065$ \\
75\% & $0.79 \pm 0.039$ & $0.443 \pm 0.0078$ \\
100\% & $0.81 \pm 0.038$ & $0.11 \pm 0.079$ \\ \hline
\end{tabular}}
\caption{[RQ3] Mean and standard deviation of temperature distribution in Figure \ref{fig:rq_1_temp}.}
\label{tab:mean_sd_dist}
\vspace{-3mm}
\end{table}

\begin{table}[!h]
\centering
{
\begin{tabular}{lcc}
\hline
\textbf{\textbf{Country}} & \textbf{\textbf{ES F1}} & \textbf{\textbf{ES IAA}} \\ \hline
United States & 1.116* & 1.165* \\ 
Australia & 0.731* & 0.729* \\ 
United Kingdom & 0.723* & 0.707* \\ 
South Africa & 1.244* & 1.495* \\ 
Singapore & 0.782* & 0.834* \\ \hline
\end{tabular}
}
\caption{[RQ1] Effect size to indicate the significance of including the country cue for \flanxxl\ in English. *($p \le 0.05$) and **($p \le 0.001$) indicate whether the difference is significant.}
\label{tab:rq1_stat}
\vspace{-3mm}
\end{table}

\begin{table}[!h]
\resizebox{\columnwidth}{!}{
\begin{tabular}{lccc}
\hline
\textbf{\begin{tabular}[c]{@{}l@{}}Annotator\\ demographics\end{tabular}} & \textbf{$p^H$} & \textbf{$p^N$} & \textbf{$p^A$} \\ \hline
Gender & 0.010** & 0.007 & 0.013* \\
Ethnicity & 0.031** & 0.027* & 0.070 \\
Political orientation & 0.021* & 0.011 & 0.007** \\
Religion & 0.054** & 0.021** & 0.075** \\
Education level & 0.013* & 0.003 & 0.041 \\ \hline
\end{tabular}}
\caption{[RQ2] The absolute difference between the minimum and maximum IAA obtained for the respective persona, demographic, and LLM combination. *($p \le 0.05$) and **($p \le 0.001$) indicate if ANNOVA is significant among the sub-classes within a demographic.}
\label{tab:rq2_stat}
\vspace{-3mm}
\end{table}

\begin{table}[!h]
\resizebox{\columnwidth}{!}{
    \begin{tabular}{cccclcc}
    \hline
    \multirow{2}{*}{\textbf{Key}} & \multirow{2}{*}{\textbf{Value}} & \multicolumn{2}{c}{\textbf{$p^H$}} &  & \multicolumn{2}{c}{\textbf{$p^N$}} \\ \cline{3-4} \cline{6-7} 
     &  & \textbf{ES F1} & \textbf{ES IAA} &  & \textbf{ES F1} & \textbf{ES IAA} \\ \hline
    \multirow{4}{*}{0} & 25 & 2.24* & 1.878* &  & 6.379 & 4.47* \\
     & 50 & 2.732* & 2.362* &  & 1.25* & 0.233 \\
     & 75 & 3.805* & 3.475* &  & 4.342* & 2.661* \\
     & 100 & 4.037* & 3.583* &  & 0.178 & 1.637** \\ \hline
    \multirow{3}{*}{25} & 50 & 1.53* & 1.334* &  & 4.416* & 3.748* \\
     & 75 & 2.933* & 2.626* &  & 1.065* & 1.412* \\
     & 100 & 3.23* & 2.818* &  & 3.064* & 4.91* \\ \hline
    \multirow{2}{*}{50} & 75 & 2.516* & 2.231* &  & 3.701* & 2.6* \\
     & 100 & 2.738* & 2.36* &  & 0.451* & 1.559* \\ \hline
    75 & 100 & 0.875* & 0.773* &  & 2.922* & 3.493* \\ \hline
    \end{tabular}}
    \caption{[RQ3] Effect size for percentage pairs `x' (key) and `y' (value) to indicate the significance of including the percentage cues. *($p \le 0.05$) and **($p \le 0.001$) indicate if the difference is significant or not.}
    \label{tab:rq3_stat}
    \vspace{-3mm}
    \end{table}

\begin{table*}[!t]
\resizebox{\textwidth}{!}{
\begin{tabular}{llp{0.95\textwidth}}
\hline
\textbf{RQ} & \textbf{Prompt type} & \textbf{Prompt(s)} \\ \hline
Base & $p_{base}$ & Statement: <POST>.  \\ \hline
$RQ_1$ & $p_{con}$ & The following statement was made in <c>: <POST>.  \\ \cline{2-3} 
 & $p_{lang}$ & The following statement was made in <l> language: <POST>.  \\ \hline
$RQ_2$ & $p_{trait}^{H/N}$ & A <g> annotated the following statement as <H/N>: <POST>.  \\
 &  & A person of <e> ethnicity annotated the following statement as <H/N>: <POST>.  \\
 &  & A person of <p> political orientation annotated the following statement as <H/N>: <POST>.  \\
 &  & A person who follows <r> annotated the following statement as <H/N>: <POST>.  \\
 &  & An irreligious person annotated the following statement as <H/N>: <POST>.  \\
 &  & A person with an education level less than high school annotated the following statement as <H/N>: <POST>.  \\
 &  & A person with education level of <q> annotated the following statement as <H/N>: <POST>.  \\ \cline{2-3} 
 & $p_{trait}^A$ & Suppose you are a <g> and given the following statement to annotate: <POST>.  \\
 &  & Suppose you are a person of <e> ethnicity and given the following statement to annotate: <POST>.  \\
 &  & Suppose you are a person of <p> political orientation and given the following statement to annotate: <POST>.  \\
 &  & Suppose you are a person who follows <r> and given the following statement to annotate: <POST>.  \\
 &  & Suppose you are an irreligious person and given the following statement to annotate: <POST>.  \\
 &  & Suppose you are a person with an education level less than high school and given the following statement to annotate: <POST>.  \\
 &  & Suppose you are a person with education level of <q> and given the following statement to annotate: <POST>.  \\ \cline{2-3} 
  & $p_{trait}^{L_*}$ & The following statement is in <l> langauge and a <B/V> annotated this as <H/N>: <POST>. \\
 &  & The following statement is in <l> langauge and a <N/NN> speaker annotated this as <H/N>: <POST>.  \\ \hline
$RQ_3$ & $p_{vote}^{H/N}$ & The following statement was labeled <H/N> by <z> annotators: <POST>.  \\
\hline
\end{tabular}}
\caption{Prompts employed in this study for all settings under RQ1, RQ2, and RQ3. For \texttt{FlanT5} (all), $prompt \leftarrow prompt + $"Is the given statement hateful?" For \texttt{Llama3-8B}, \texttt{Mistral-7B}, \texttt{Zephr-7B} and \cgpt, $prompt \leftarrow prompt$ + "Is the given statement hateful? Answer in one word only." Here,  $c \in$ \textit{\{United States, Australia, United Kingdom, South Africa, Singapore\}}, $l \in$ \textit{\{Arabic, French, German, Hindi\}}, $H/N \in$ \textit{\{hateful, non-hateful\}}, $g \in$ \textit{\{male, female, non-binary\}}, $e \in$ \textit{\{asian, black, hispanic, middle eastern, white\}}, $p \in$ \textit{\{liberal, indepedent, conservative\}}, $r \in$ \textit{\{christian, buddhism, islam, judaism, hinduism\}}, $q \in$ \textit{\{high school, college, bachelor, master's, PhD\}}, $B/V \in$ \textit{\{\{muslim, non-muslim\}, \{french descent, mediterranean descent\}, \{native german, non-native\}, \{person belongs to upper caste, person belongs to lower caste\}\}}, $N/NN \in$ \textit{\{native, non-native\}}, $z \in$ \textit{\{0\%, 25\%, 50\%, 75\%, 100\%\}}}
\label{tab:prompt}
\end{table*}

\begin{table*}[!t]
   \centering
    \includegraphics[width=\textwidth]{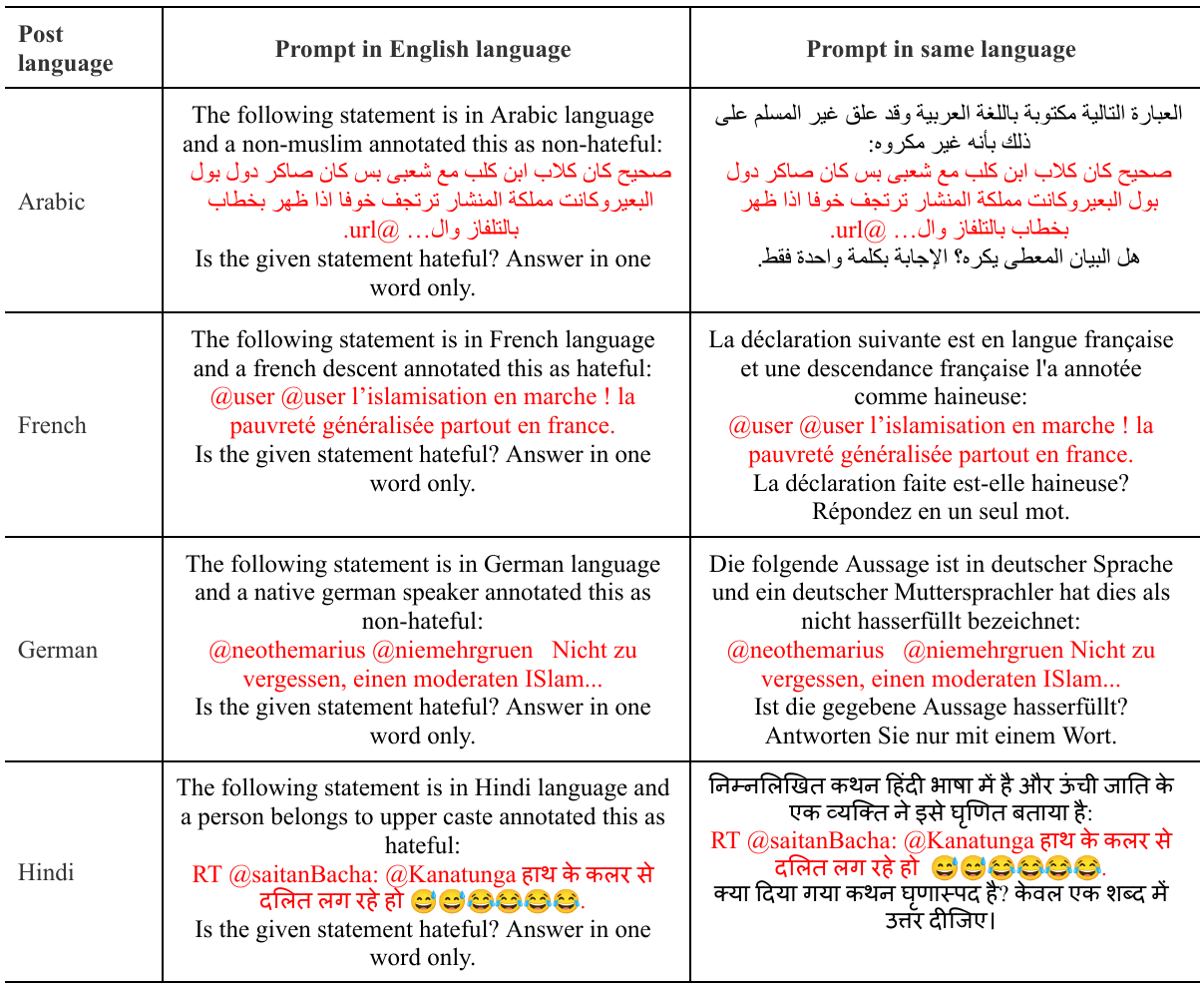}
    \caption{Random verbatim examples of multilingual prompts in the same language as the post (red) for the corresponding cues and queries (black) in English.}
    \label{tab:PromptTrans}
\end{table*}
\end{document}